\documentclass[twoside,11pt]{article}

\usepackage{blindtext}

\usepackage{jmlr2e}
\usepackage[utf8]{inputenc}
\usepackage{amsmath,amssymb}
\usepackage{algorithm}
\usepackage{algpseudocode}
\usepackage{enumitem}

\usepackage{lastpage}

\ShortHeadings{AutoGMM}{}
\firstpageno{1}

\begin{document}

\title{AutoGMM: Automatic Gaussian Mixture Modeling in Python}


\ShortHeadings{AutoGMM}{Liu, Athey, Pedigo, and Vogelstein}
\firstpageno{1}

\author{\hspace{-1.5mm}
  \name Tingshan Liu\thanks{Equal contribution.} \email tliu68@jhu.edu \\
  \addr Johns Hopkins University
  \AND
\name Thomas L.~Athey\footnotemark[1] \email tom.l.athey@gmail.com \\
  \addr Massachusetts Institute of Technology
  \AND
  \name Benjamin D.~Pedigo \email ben.pedigo@alleninstitute.org \\
  \addr Johns Hopkins University
  \AND
  \name Joshua T.~Vogelstein \email jovo@jhu.edu \\
  \addr Johns Hopkins University
}

\editor{}

\maketitle

\vspace{-15pt}\begin{abstract}

The exponential growth of complex data demands fully automatic clustering. Gaussian mixture models (GMMs) provide uncertainty-aware grouping but often require expertise to specify hyperparameters, e.g., component count and covariance structure. While \texttt{mclust} (R) automates this via Bayesian Information Criterion (BIC), Python lacks a comparable tool. We introduce \textsc{AutoGMM}, an open-source Python package automating GMM via strategic initialization using an agglomerative Mahalanobis heuristic, and parallelized model selection by information criteria. \textsc{AutoGMM} is a drop-in tool that yields strong out-of-the-box performance on classic benchmarks, targeted stress tests, and two real datasets, with favorable runtime scaling. The code is available at \url{https://github.com/neurodata/AutoGMM} with tests and reproducible workflows.

\end{abstract}


\section{Introduction}
Clustering is a foundational operation in data analysis, supporting applications from neuroscience \citep{yeo2011organization, schaefer2018local}, and precision medicine \citep{perou2000molecular, cancer2011integrated} to large-scale identity resolution in computer vision \citep{nguyen2021clusformer}. As datasets grow in size and complexity, the demand for accessible, reliable, and reproducible tools that reveal structure without bespoke tuning increases. A particularly promising path is model-based clustering with Gaussian mixture models (GMMs), which yield interpretable, uncertainty-aware partitions together with a likelihood for downstream inference and principled model comparison. However, in practice GMM performance is highly sensitive to initialization, the unknown number of components, and covariance regularization. Moreover,  finite-sample covariance estimates can be ill-conditioned, complicating estimation, allowing degeneracies, and undermining reproducibility. Although the R ecosystem’s \texttt{mclust} established a gold standard via hierarchical initialization and Bayesian Information Criterion (BIC)-driven model selection \citep{fraley2002model, scrucca2023model}, the Python ecosystem lacks an equally robust, automated counterpart, creating a persistent gap for practitioners working in Python-dominated workflows.

We present \textsc{AutoGMM}, an open-source Python package that automates GMM fitting end-to-end with a scikit-learn-compatible API. The system automates initialization, stabilizes covariance estimation, and performs information-criterion model selection with an optional spectral front-end. Experiments show strong out-of-the-box performance on classic benchmarks, targeted low and high-dimensional stress tests, and two real datasets, with favorable runtime scaling.

\vspace{-5pt}
\section{Method}
\vspace{-2pt}
Given data $X\in\mathbb{R}^{n\times d}$, a component range $[K_{\min}, K_{\max}]$, and covariance classes $\mathcal{C}\subseteq$
\{\texttt{spherical}, \texttt{diag}, \texttt{tied}, \texttt{full}\} (as in \texttt{sklearn.mixture.GaussianMixture}), \textsc{AutoGMM} returns the model
minimizing an information criterion (IC), Bayesian Information Criterion (BIC) by default, with Akaike Information Criterion (AIC) available. 
The pipeline has three stages:
\begin{enumerate}[leftmargin=12pt,itemsep=0.2pt,topsep=0.2pt]
\item \emph{Initialization (three seeds).} For each $K\in[K_{\min}, K_{\max}]$, candidate labelings come from (i) K-means with multiple restarts; (ii) Ward-Euclidean agglomeration cut at $K$; and (iii) Ward-Mahalanobis, which replaces Euclidean distances by a pooled-Mahalanobis variant that measures distances using a pooled-precision $\widehat{\Sigma}^{-1}$ estimated after PCA with OAS shrinkage. These candidates seed the expectation-maximization (EM) routine.
\vspace{-6pt}\item \emph{EM with regularization.} For each initializer and covariance class $c\in\mathcal{C}$, we fit a GMM by EM. After updating covariances in the M-step, we stabilize them by (i) adding a small diagonal ridge and (ii) flooring eigenvalues which guarantees $\widehat{\Sigma}\succ 0$ and bounds condition numbers. Regularization hyperparameters are fixed and require no tuning.
\vspace{-6pt}\item \emph{Selection and parallelism.} For every $(K,c)$ and initializer, we compute the chosen IC and select the minimum. The grid over $(K,c)$ is evaluated in parallel.
\end{enumerate}

\texttt{mclust} uses a model-based hierarchical clustering (MBHC) approach to initialize the EM algorithm that they use.  Because \texttt{scikit-learn}'s agglomerative clustering is not model based, we introduce a Mahalanobis–Ward agglomerative variant that uses pooled regularized precision to approximate MBHC.

We also evaluate an optional spectral front-end and retain the better (raw vs embedded) IC. We build a symmetrized $k$-NN graph (default $k=\lfloor\sqrt{n}\rfloor$, tunable \texttt{n\_neighbors}) in Euclidean distance. Edge weights are defined by a Gaussian (RBF) kernel $w_{ij}=\exp{(-\gamma\|x_i-x_j\|_2^2)}$. $\gamma$ is the so-called  global median heuristic;  letting $\sigma$ be the median of all pairwise Euclidean distances from the samples, then $\gamma=1/(2\sigma^2)$. To ensure a connected graph, we add the edges of the minimum spanning tree computed form the same Euclidean distances. We then compute adjacency/Laplacian spectral embedding where the embedding dimension is chosen by a maximum-likelihood criterion under probabilistic PCA \citep{minka2000automatic}. See Appendix \ref{sec:theory} for details.

\vspace{-10pt}
\section{Evaluation}
We evaluate \textsc{AutoGMM} on synthetic and real data against strong baselines. Unless noted, results aggregate 50 trials with fixed seeds and we report the Adjusted Rand Index (ARI; $1=$ perfect agreement, $0\approx$ chance). Model selection defaults to BIC; the number of components and covariance class are selected automatically. All experiments are reproducible via scripts in the repository.

\begin{figure}[ht!]
    \centering
    \includegraphics[width=\linewidth]{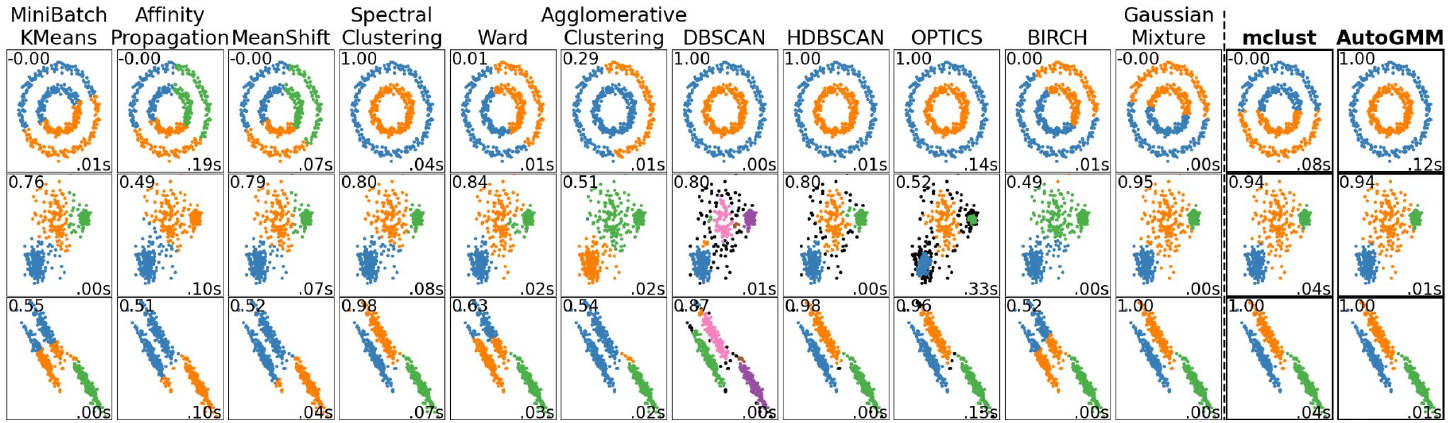}\vspace{70pt}
    \vspace{-90pt}\caption{\textbf{Classic benchmarks.} Rows: datasets; columns: methods. Each panel shows ARI (top-left) and runtime in seconds (bottom-right). \textsc{mclust} and \textsc{AutoGMM} automatically choose both $K$ and covariance via BIC. All other baselines do not perform model selection. They are run with $K$ fixed to ground truth and default hyperparameters per the \texttt{scikit-learn} recipe. In particular, \texttt{GaussianMixture} uses the default \texttt{covariance\_type}=\texttt{full} and $k$-means initialization. See the full gallery in Figure \ref{fig:clus_comp_app}.}\vspace{-10pt}
    \label{fig:clus_comp}
\end{figure}

First, we replicate the standard \textsc{scikit-learn} clustering gallery and compare \textsc{AutoGMM} to representative baselines including \texttt{mclust} (Figures \ref{fig:clus_comp}, \ref{fig:clus_comp_app}). The optional spectral ($k$-NN-RBF) front-end yields perfect ARI on nonconvex shapes (circles, moons). On blob-like data, the default (non-kernel) configuration matches the strongest methods while remaining competitive in runtime.

\begin{figure}[ht!]
    \centering
    \includegraphics[width=0.48\linewidth]{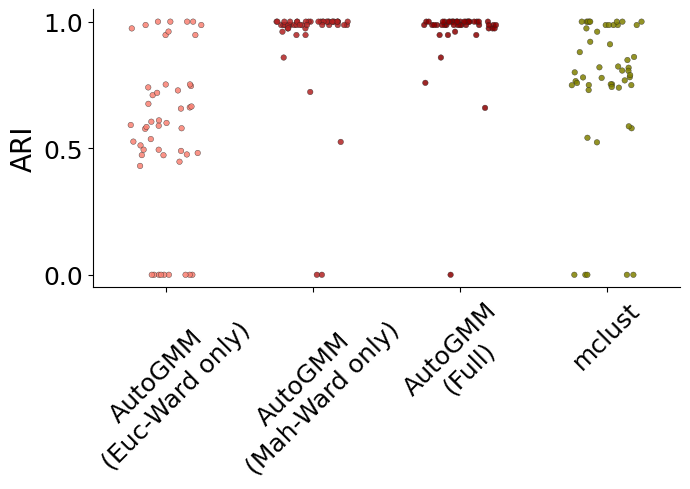}
    \includegraphics[width=0.48\linewidth]{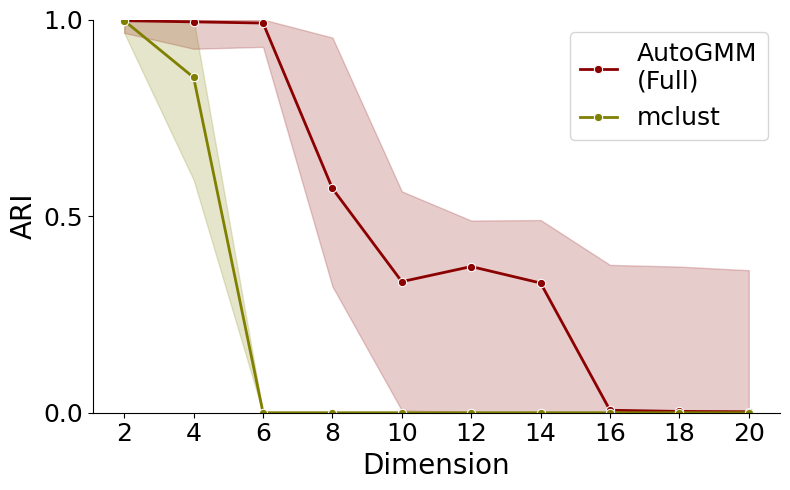}
    \vspace{-10pt}\caption{\textbf{Stress-test ablations (left) and high-dimensional scaling (right).} Left: ARIs on an anisotropic stress test. Right: Median ARI versus dimension when the benchmark anisotropic dataset is lifted to higher dimensions; shaded bands show the IQR range across runs.}\vspace{-5pt}
    \label{fig:high_dim}
\end{figure}

To isolate robustness to covariance ill-conditioning, we construct mixtures of elongated Gaussians and increase dimension $d$. We concatenate independent replicas of the same 2D anisotropic mixture: each replica was produced with a different seed and the same linear transformation so that geometry is preserved while $d \in \{2, 4, \dots, 20\}$. We perform an initialization ablation: for each 
$d$, we run \textsc{AutoGMM} three times—using only Ward–Euclidean seeds, only Ward–Mahalanobis seeds, all initializations, respectively. Figure~\ref{fig:high_dim} reports median ARI with interquartile bands. Ward-Mahalanobis and \textsc{AutoGMM} (full) achieve near-perfect ARI while Ward-Euclidean and \texttt{mclust} occasionally fail. For the high-dimensional lift, as $d$ increases, \textsc{AutoGMM} degrades gracefully and remains above \texttt{mclust} which drops sharply on this stress test.

\begin{figure}[ht!]
    \centering
    \includegraphics[width=\linewidth]{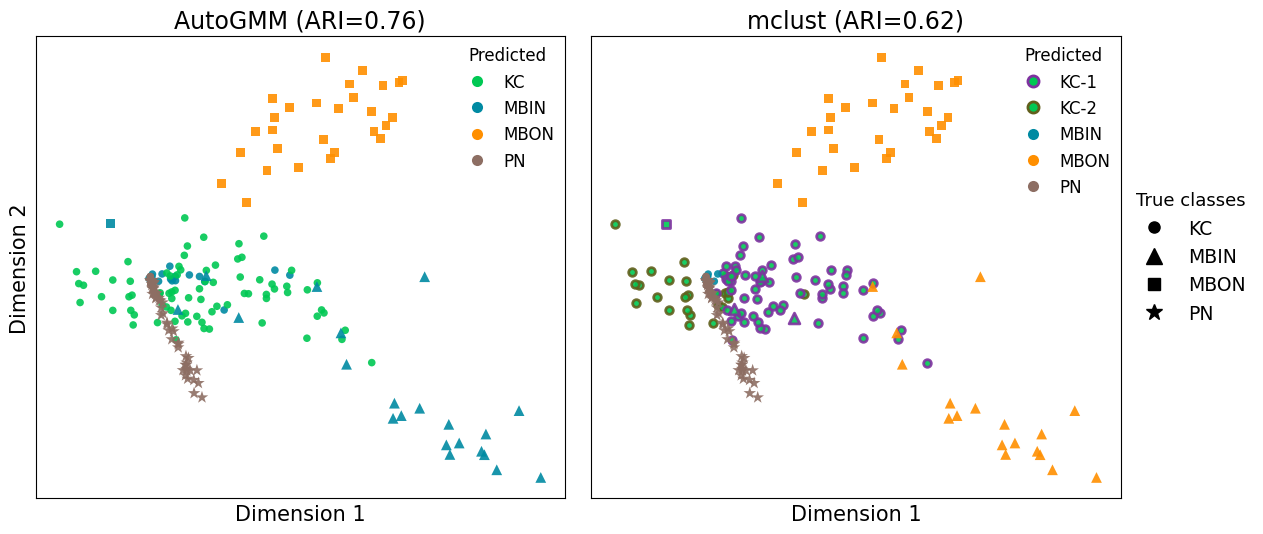}
    \vspace{-20pt}\caption{\textbf{ \textit{Drosophila} connectome.} Mushroom body data (6-D, $n$=213) were embedded by ASE and clustered with \textsc{AutoGMM} (left) and \texttt{mclust} (right). Cell types are considered as true classes. Predicted cluster labels are aligned to truth by solving a linear assignment on the cluster-class contingency matrix with the Hungarian method to maximize overlap \citep{kuhn1955hungarian}.  The top two dimensions are plotted.}\vspace{-5pt}
    \label{fig:real_droso}
\end{figure}

We include two real datasets to illustrate practical behavior. On a labeled Drosophila mushroom body connectome with four principal cell types \citep{priebe2017semiparametric}, \textsc{AutoGMM} attains 0.76 versus 0.62 for \texttt{mclust} (Figure \ref{fig:real_droso}). On a cancer/normal fragmentomics cohort with binary labels \citep{curtis2025fragmentation}, \textsc{AutoGMM} achieves 0.30 versus 0.25 for \texttt{mclust} (Figure \ref{fig:real_cancer}). In both cases $K$ and covariance structure are selected automatically by BIC, and preprocessing is identical across methods.

Finally, we study runtime as a function of sample size  $n$ and dimension $d$ using an isotropic Gaussian synthetic dataset. We report medians with interquartile bands over repeated runs on a single workstation (Fig \ref{fig:runtime}). When sweeping sample size at fixed $d=10$, \textsc{AutoGMM} is faster than \texttt{mclust} for moderate $n$. When sweeping dimension at fixed $n=1000$, \textsc{AutoGMM} remains consistently faster.

\vspace{-0.1cm}
\section{Discussion}
\textsc{Autogmm} addresses a practical gap in the Python ecosystem: a  reproducible tool that makes Gaussian mixture modeling dependable without bespoke pipelines. Empirically, AutoGMM matches or exceeds strong baselines on classic galleries, remains robust under high-dimensional anisotropy, and improves ARI on two real datasets. However, limitations remain and suggest focused extensions. Ward-based seeding is $\mathcal{O}(n^2)$ in time/memory, and pooled-precision Mahalanobis seeding can be suboptimal when component covariances differ markedly. Future directions include scalable agglomeration, and more robust graph construction with unsupervised geodesic forest kernels \citep{madhyastha2020geodesic}.


\clearpage
\acks{We acknowledge support from the National Science Foundation (Grant No. 2014862). We also thank Dr. Carey Priebe for helpful discussions.}







\vskip 0.2in
\bibliography{sample}

\appendix
\section{Figures}
\setcounter{figure}{0}
\renewcommand{\thefigure}{A.\arabic{figure}}

\begin{figure}[ht!]
    \centering
    \includegraphics[width=\linewidth]{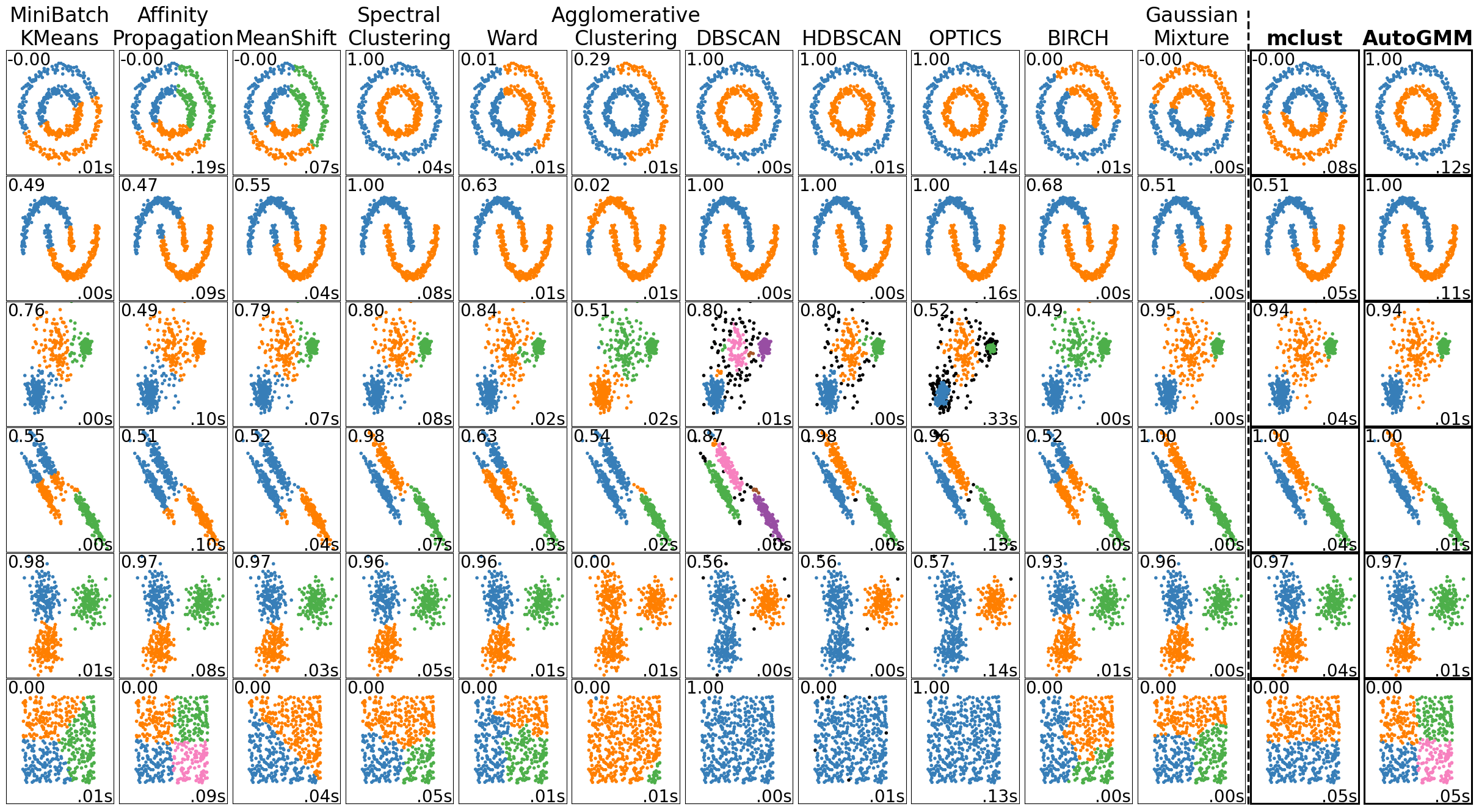}
    \caption{\textbf{Full classic benchmarks.} Extended \textsc{scikit-learn} gallery with the same protocol as Figure \ref{fig:clus_comp}.}
    \label{fig:clus_comp_app}
\end{figure}

\begin{figure}[ht!]
    \centering
    \includegraphics[width=0.8\linewidth]{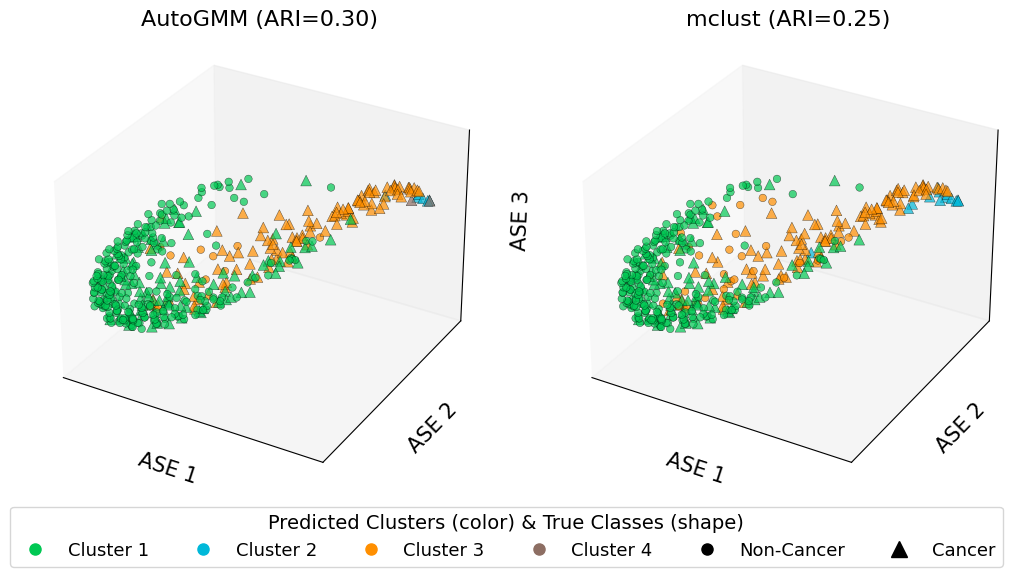}
    \caption{\textbf{Real cancer data.} The fragmentomics data (12-dimensional, 466 samples) was embedded by ASE and clustered with \textsc{AutoGMM} (left) and \texttt{mclust} (right). Cancer status is considered as true class labels.}
    \label{fig:real_cancer}
\end{figure}

\begin{figure}[ht!]
    \centering
    \includegraphics[width=0.75\linewidth]{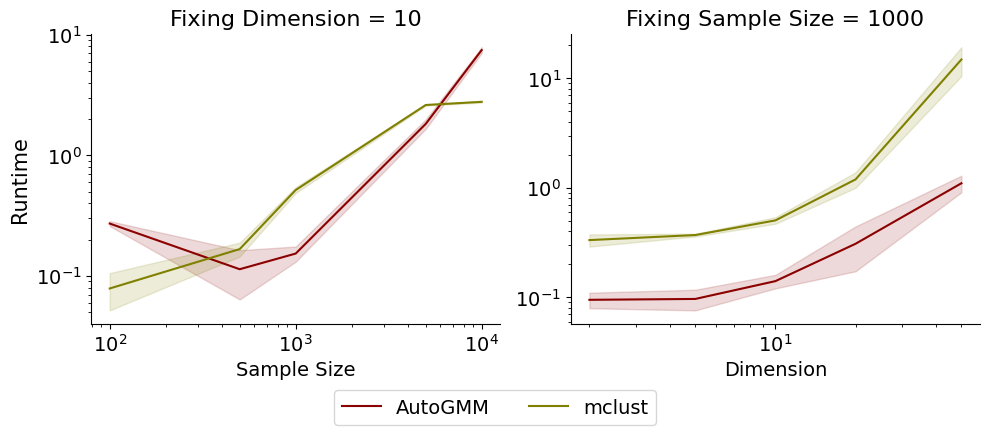}
    \caption{\textbf{Runtime scalability.} Mean wall-clock time with $\pm$1 standard deviation bands.}
    \label{fig:runtime}
\end{figure}

\section{Theoretical Remarks}\label{sec:theory}
\paragraph{Notation.}
For a sample covariance estimate $\widehat{\Sigma}\in\mathbb{R}^{d\times d}$ with eigendecomposition
$\widehat{\Sigma}=U\Lambda U^\top$, let $\Lambda=\mathrm{diag}(\lambda_1,\ldots,\lambda_d)$ and
define the \emph{eigenvalue floor} at level $\varepsilon>0$ by
\[
\widehat{\Sigma}_{\mathrm{floor}} \;=\; U\,\mathrm{diag}\!\big(\max\{\lambda_j,\varepsilon\}\big)_{j=1}^d\,U^\top.
\]
For comparison, a common ridge shrinkage is $\widehat{\Sigma}_{\mathrm{ridge}}=\widehat{\Sigma}+\varepsilon I_d$. We write $\kappa_2(M)=\lambda_{\max}(M)/\lambda_{\min}(M)$ for the spectral condition number. For \texttt{diag} (resp. \texttt{spherical}) covariances, flooring reduces to $\sigma_j^2\leftarrow \max\{\sigma_j^2,\varepsilon\}$ (resp. $\sigma^2\leftarrow \max\{\sigma^2,\varepsilon\}$).

\subsection{Eigenvalue Thresholding}
\paragraph{Motivation.}
In finite samples, especially with anisotropy or $n$ not much larger than $d$, component covariances in GMMs can become ill-conditioned or singular, leading to numerical failures and the well-known “likelihood blow-up” where a component collapses onto a few points \citep{fraley2002model}.

\paragraph{Remark (conditioning and stability).}
The flooring operation guarantees positive definiteness and controls the condition number:
$$\lambda_{\min}(\widehat{\Sigma}_{\mathrm{floor}})\ge \varepsilon \text{\, and \,}
\kappa_2(\widehat{\Sigma}_{\mathrm{floor}})\le \lambda_{\max}(\widehat{\Sigma})/\varepsilon.$$
Consequently, $\|\widehat{\Sigma}_{\mathrm{floor}}^{-1}\|_2=1/\lambda_{\min}(\hat\Sigma_{\mathrm{floor}}) \le 1/\varepsilon$, so all Mahalanobis distances and EM updates remain well-defined and numerically stable. 

\paragraph{Remark (connection to penalization).}
The covariance M-step for a Gaussian with the spectral constraint $\lambda_{\min}(\Sigma)\ge \varepsilon$ has the closed-form solution $\widehat\Sigma_{\mathrm{floor}}$ obtained by flooring the eigenvalues of
the sample covariance. A common alternative is \emph{ridge shrinkage}, which replaces $\widehat{\Sigma}$ by $\widehat{\Sigma}+\varepsilon I_d$. Both enforce $\Sigma\succ 0$. preserves
principal directions and raises only small eigenvalues, whereas ridge shifts all eigenvalues by the same amount. In our implementation, flooring is applied to each (tied/full) covariance after the M-step, which prevents degenerate components and improves EM robustness.

\subsection{Mahalanobis Initialization}
\paragraph{Motivation.}
Ward-Euclidean agglomeration minimizes the increase in within-cluster sum of squares and is optimal when clusters are spherical with equal variance. For anisotropic clusters, a Mahalanobis metric aligns the merge rule with the underlying Gaussian geometry.


\subsection{Kernel Embedding}
\paragraph{Motivation.}
Nonconvex cluster shapes (e.g., circles, moons) violate the Euclidean mixture assumption in the ambient space. Spectral embeddings map data into a space where clusters are more nearly Euclidean-separable, enabling effective downstream Gaussian modeling.

\paragraph{Remark (ASE/LSE as Euclidean latent positions).}
Let $A$ be the binary $k$-NN adjacency (after MST injection) and $W$ the RBF-weighted graph. ASE is computed from top eigenpairs of $A$; LSE uses the top eigenpairs of the normalized Laplacian $L=D^{-1/2}WD^{-1/2}$. Under standard random-graph models (e.g., SBM and degree-corrected SBM), ASE/LSE recovers latent positions up to an orthogonal transform, and those positions concentrate around cluster-specific means \citep{sussman2012consistent,tang2018limit}. Empirically, this “Euclideanizes’’ nonconvex geometry so that a GMM (with BIC model selection) is appropriate. Our engineering additions ($k$-NN graph with MST injection and RBF weights; MLE-based target dimension) improve connectivity and finite-sample stability without changing this rationale.

\end{document}